\documentclass{article}

\usepackage[margin=1in]{geometry}
\usepackage[slantedGreek]{mathpazo}
\usepackage[round]{natbib}
\usepackage{graphicx,amsmath,amsfonts,dsfont, adjustbox}
\usepackage{float}
\usepackage{tikz}

\title{Agri-GNN: A Novel Genotypic-Topological Graph Neural Network Framework Built on GraphSAGE for Optimized Yield Prediction}
\author{	
  \makebox[.4\linewidth]{Aditya Gupta}\\\textit{\small  William Fremd High School}\\\textit{\small  Palatine, IL}\\\and
  \makebox[.4\linewidth]{Asheesh Singh}\\\textit{\small  Agronomy Department}\\\textit{\small  Iowa State University}
}

\graphicspath{{fig/}}

\begin{document}

\maketitle

\emph{The ultimate goal of farming is not the growing of the crops, but the cultivation and perfection of human beings.   \hspace{1.0in} ---Masanobu Fukuoka}

\vspace{0.1in}
\begin{abstract}
Agriculture, as the cornerstone of human civilization, constantly seeks to integrate technology for enhanced productivity and sustainability. This paper introduces \textit{Agri-GNN}, a novel Genotypic-Topological Graph Neural Network Framework tailored to capture the intricate spatial and genotypic interactions of crops, paving the way for optimized predictions of harvest yields. \textit{Agri-GNN} constructs a Graph $\mathcal{G}$ that considers farming plots as nodes, and then methodically constructs edges between nodes based on spatial and genotypic similarity, allowing for the aggregation of node information thorugh a genotypic-topological filter. Graph Neural Networks (GNN), by design, consider the relationships between data points, enabling them to efficiently model the interconnected agricultural ecosystem. By harnessing the power of GNNs, \textit{Agri-GNN} encapsulates both local and global information from plants, considering their inherent connections based on spatial proximity and shared genotypes, allowing stronger predictions to be made than traditional Machine Learning architectures. \textit{Agri-GNN} is built from the GraphSAGE architecture, because of its optimal calibration with large graphs, like those of farming plots and breeding experiments. \textit{Agri-GNN} experiments, conducted on a comprehensive dataset of vegetation indices, time, genotype information, and location data, demonstrate that \textit{Agri-GNN} achieves an $R^2 = .876$ in yield predictions for farming fields in Iowa. The results show significant improvement over the baselines and other work in the field. \textit{Agri-GNN} represents a blueprint for using advanced graph-based neural architectures to predict crop yield, providing significant improvements over baselines in the field.
\end{abstract}


\vspace{0.1in}

\section{Introduction}

In an era characterized by escalating climate change, which is resulting in unpredictable weather patterns and increasing environmental stresses, the agricultural sector faces significant challenges~\citep{anwar2013adapting}. Unforeseen climatic events such as droughts, floods, and extreme temperatures are impacting crop yields, highlighting the imperative for advanced, precise, and resilient crop yield prediction models~\citep{kuwayama2019estimating}. Amidst this backdrop of climatic uncertainties~\citep{shrestha2012climate}, the necessity for accurate and comprehensive crop yield predictions is more acute than ever. A robust system that can efficiently integrate diverse data types and provide a detailed and holistic understanding of the agricultural ecosystem is crucial for mitigating the impacts of climate change on agriculture.

The agricultural ecosystem is inherently complex and interconnected, with numerous factors playing a pivotal role in determining crop yields. Traditional Machine Learning models, while powerful, often fall short in effectively capturing these intricate relationships, as they generally treat data points as independent entities~\citep{liu2020towards}. The limited capacity of these models to handle relational data and their inability to seamlessly integrate diverse data types such as spatial, temporal, and genetic information, hamper their effectiveness in providing comprehensive and accurate crop yield predictions. Moreover, these models tend to be data-hungry, requiring substantial labeled data for training, which is often a significant challenge in the agricultural context~\citep{majumdar2017analysis}. On the contrary, Graph Neural Networks (GNNs) stand out as a more apt choice for this scenario. GNNs, by design, consider the relationships between data points, enabling them to efficiently model the interconnected agricultural ecosystem. They can effectively synthesize diverse data types into a unified framework, offering a more holistic and nuanced understanding of the factors influencing crop yields~\citep{zhou2020graph}. The ability of GNNs to work with limited labeled data and their flexibility in handling various data modalities make them a superior choice for developing robust and resilient crop yield prediction models in the face of climate change.

In light of this, the present study introduces Agri-GNN, a pioneering approach employing GNNs to offer an inclusive representation of the agricultural ecosystem.  \textit{Agri-GNN} considers farming plots as nodes, and then constructs edges between nodes based on spatial and genotypic similarity, allowing for the aggregation of node information from a refined selection of nodes. This allows for the model to refine the noise that exists in the dataset and focus yield prediction efforts for each node on the most similar nodes in terms of genotypic and spatial similarity. \textit{Agri-GNN} stands out with its capacity to amalgamate diverse data modalities into a cohesive framework, adeptly capturing the complex interaction among genetic, environmental, and spatial factors~\citep{meng2018efficient}. This innovative model transcends traditional methodologies by conceptualizing the agricultural ecosystem as a connected network, where each crop, viewed as an active node, is influenced by its immediate environment and genetic context.

The employment of the GraphSAGE architecture~\citep{hamilton2017inductive}, significantly bolsters the effectiveness of \textit{Agri-GNN}. GraphSAGE is known for its inductive learning approach, where it leverages node attribute information to generate representations for data not seen during the training process. This approach is particularly beneficial for the extensive and heterogeneous datasets that are commonplace in the field of agriculture. Traditional machine learning models often struggle with such diverse and expansive data, leading to suboptimal predictions and insights. However, the GraphSAGE architecture, with its innovative inductive learning, excels in processing and learning from large datasets, thereby ensuring the robustness of \textit{Agri-GNN}.

In agriculture, datasets encompass a wide range of information including weather conditions, soil types, and genetic information, the ability to effectively handle and learn from such data is crucial for accurate yield predictions. The GraphSAGE architecture equips \textit{Agri-GNN} with this capability, allowing it to seamlessly integrate and process diverse data types to generate detailed and reliable yield predictions. This level of granular insight is important for making informed decisions in agricultural planning and management, ultimately contributing to enhanced productivity and sustainability.

By using the GraphSAGE architecture, \textit{Agri-GNN} is not just limited to data seen during training. It can generalize and adapt to new data, ensuring that the model remains relevant and useful as more agricultural data becomes available. This adaptability is essential in the dynamic field of agriculture, where new data and insights continuously emerge. The advanced architecture thereby not only enhances \textit{Agri-GNN}'s predictive accuracy but also bolsters its longevity and relevance in the agricultural sector, making it a valuable tool for tackling the challenges of modern agriculture. \textit{Agri-GNN}'s modular and scalable design ensures its adaptability to the fast-paced evolution of the agricultural sector. This flexibility allows for the effortless integration of emerging data sources and insights, ensuring the model remains relevant and effective in a changing landscape~\citep{gandhi2016review}.

\textit{Agri-GNN} embodies a transformative shift that is taking place in agricultural modeling, providing a novel perspective that comprehensively addresses the complexity and interconnectedness of farming systems. By offering a nuanced, data-driven lens, \textit{Agri-GNN} stands as a robust tool for navigating the multifaceted challenges of modern agriculture, particularly in the context of a changing climate.

\section{Literature Review}\label{sec:lit}

Plant breeding specialists are focused on discovering high-quality genetic variations that fulfill the needs of farmers, the wider agricultural sector, and end consumers. One of the key traits often scrutinized is seed yield, particularly in row crops~\citep{singh2021plant}. Traditional ways of assessing seed yield involve the laborious and time-restricted activity of machine-harvesting numerous plots when the growing season concludes. This data then informs decisions about which genetic lines to either advance or discontinue in breeding programs. This approach is highly resource-intensive, requiring the harvesting of thousands of test plots each year, thus presenting operational challenges.

In response to these issues, advancements in technology are being harnessed to develop more efficient alternatives. An increasing number of scientists and plant breeders are adopting the use of remote sensing technology, integrated with machine learning methods. This enables more timely predictions of seed yield, substantially cutting down on labor and time requirements during the crucial harvest phase~\citep{zongpengetal22,yoosefzadeh2021genome,marianaetal21,shooketal21,riera21,guo21,singh21}.

The newly introduced Cyber-Agricultural System (CAS) takes advantage of cutting-edge continual sensing technology, artificial intelligence, and smart actuators for enhancing both breeding and production in agriculture~\citet{CAS}. Integral to CAS is the concept of phenotyping, which employs sophisticated imaging and computational techniques to streamline the gathering and interpretation of data, thereby facilitating better yield forecasts~\citet{singh2021high}. Numerous studies have honed in on high-throughput phenotyping through the use of drones, investigating yield predictions in crops such as cotton, maize, soybean, and wheat~\citet{herr2023unoccupied}. Beyond the 2D data collected by drones, research has demonstrated the value of canopy fingerprints, which offer unique insights into the 3D structure of soybean canopies via point cloud data~\citet{young2023canopy}. Despite these advances, there is still scope for refining models that amalgamate diverse datasets, including but not limited to soil features and hyperspectral reflectance, for a more holistic grasp of soybean yields. The soil's physical and chemical attributes play a crucial role in nutrient availability, thereby impacting plant health and growth. Incorporating these soil characteristics could potentially enhance the precision of yield prediction models.

In recent years, the application of neural networks in crop yield prediction has moved beyond traditional architectures to more complex models like Convolutional Neural Networks (CNNs) and Recurrent Neural Networks (RNNs) \citep{dahikar2014agricultural,gandhi2016rice}. 

Much work in the field takes advantage of remote sensing data, such as satellite images or NDVI, for yield predictions \citep{you2017deep,nevavuori2019crop,kim2019comparison}. While these methods have shown promise, they often struggle to capture the direct relationships between environmental factors and crop yields. Previous research has also focused on using environmental factors like temperature and rainfall directly as inputs for yield prediction models \citep{ccakir2014yield,khaki2019crop}, but the same failure to capture direct relationships accurately has been seen.

Against this backdrop, Graph Neural Networks (GNNs) offer a significant advancement by incorporating spatial relationships or neighborhood information into the prediction models. By adding the spatial context through GNNs, recent models provide a more nuanced understanding of how localized factors can impact yield, enhancing prediction accuracy~\citep{fan2022gnn, park2019physics, sajitha2023smart}. Particularly, \cite{fan2022gnn} uses a GNN-RNN based approach that shows promising results, but fails to generalize on graphs of large size. The GraphSAGE model \citep{hamilton2017inductive} stands out as a notable innovation in leveraging spatial and neighborhood information for more accurate crop yield predictions. The incorporation of GraphSAGE allows us to effectively capture localized contextual information, thereby refining our understanding of how specific factors in localized areas can influence crop yields. This results in an enhanced level of accuracy in our yield predictions.

Graph Neural Networks offer a promising avenue for enhancing the state-of-the-art in crop yield prediction. They facilitate the integration of various types of data and have the potential to significantly improve the accuracy of existing models. Future research should focus on leveraging these capabilities to build models that can generalize well across different conditions and scales.
\section{Background}\label{sec:background}
In this section, we provide background information on Graph Neural Networks and GraphSAGE. We also provide a background on the data features used in the creation of Agri-GNN. This background information is necessary for understanding Agri-GNN.
\subsection{Graph Neural Networks}

Graphs are a robust and versatile means of capturing relationships among entities, known as nodes, and their interconnections, termed edges. Graph Neural Networks elevate this representational power by extending classical neural network architectures to operate directly on graph-structured data. These networks are particularly effective at generating meaningful node-level or graph-level embeddings, representations that can be subsequently used in various downstream tasks such as node classification, link prediction, and community detection. For an exhaustive review of the techniques and methods employed in GNNs, we direct the reader to seminal survey papers by \cite{battaglia2018relational} and \cite{chami2021horopca}.

A GNN can be represented as \( f(A, X; W) \rightarrow y \), where \( y \) denotes the set of predicted outcomes (e.g., plot yield predictions), \( A \) is an \( n \times n \) adjacency matrix encapsulating the graph structure, \( X \) is an \( n \times p \) feature matrix detailing node attributes, and \( W \) represents the trainable weight parameters of the network. In this function, \( A \) and \( X \) serve as inputs, and the network \( f \) is parameterized by \( W \).

One of the distinguishing characteristics of GNNs lies in their ability to accumulate and propagate information across nodes~\citep{hamilton2017inductive}. A node's feature vector is updated iteratively based on the feature vectors of its neighboring nodes. The depth or number of layers, \( \ell \), in the GNN controls which neighbors are involved in these updates. Specifically, the final representation of a node only includes information from nodes that are at most \( \ell \)-hop distance away. This scope of nodes and edges involved in the computation of a node's representation is termed its computational graph, formally defined as \( G = (A, V) \). Here, \( A \) is the adjacency matrix for the subgraph, and \( X_v \) is the feature matrix for nodes within the \( \ell \)-hop neighborhood of the node \( v \).

\subsection{GraphSAGE}

GraphSAGE (Graph Sample and Aggregation) is a pioneering extension to general Graph Neural Networks and was specifically designed to address challenges such as scalability and inductive learning on large graphs ~\citep{hamilton2017inductive}. Unlike traditional GNN architectures that require the entire graph to be loaded into memory for training, GraphSAGE leverages a sampling strategy to extract localized subgraphs, thereby allowing for mini-batch training on large-scale graphs. The key innovation in GraphSAGE is its novel aggregation mechanism, which uses parameterized functions to aggregate information from a node's neighbors. These functions can be as simple as taking an average or as complex as employing a neural network for the aggregation process.

GraphSAGE is expressed as \( f_{\text{SAGE}}(A, X; W) \rightarrow y \), where \( f_{\text{SAGE}} \) represents the GraphSAGE model, \( y \) is the output (such as node embeddings or graph-level predictions), \( A \) is the adjacency matrix, \( X \) is the feature matrix, and \( W \) are the trainable parameters. Like generic GNNs, GraphSAGE accumulates and combines information from a node's neighborhood to update its feature representation. However, GraphSAGE can generalize to unseen nodes during inference by leveraging learned aggregation functions, making it particularly valuable for evolving graphs where the node set can change over time. It has been employed effectively in diverse applications such as social network analysis, recommendation systems, and even in specialized fields like computational biology and agronomy, showcasing its adaptability and efficiency\citep{xiao2019social}.

In formal terms, the \( l \)-th layer of GraphSAGE is defined. The aggregated embedding from neighboring counties, denoted \( \mathbf{a}_{c,t}^{(l)} \), is calculated using the function \( g_l \) applied to the embeddings \( \mathbf{z}_{c',t}^{(l-1)} \) for all neighboring counties \( c' \) of county \( c \), represented as:

\[
\mathbf{a}_{c,t}^{(l)} = g_l(\{\mathbf{z}_{c',t}^{(l-1)},\forall c'\in\mathcal N(c)\})
\]

Here, \( \mathcal N(c)=\{c', \forall A_{c,c'}=1\} \) denotes the set of neighboring counties for \( c \).

The embedding for the \( l \)-th layer, \( \mathbf{z}_{c,t}^{(l)} \), is then obtained by applying a non-linear function \( \sigma \) to the product of a weight matrix \( \mathbf{W}^{(l)} \) and the concatenation of the last layer's embedding \( \mathbf{z}_{c,t}^{(l-1)} \) and \( \mathbf{a}_{c,t}^{(l)} \):

\[
\mathbf{z}_{c,t}^{(l)} = \sigma(\mathbf{W}^{(l)}\cdot (\mathbf{z}_{c,t}^{(l-1)}, \mathbf{a}_{c,t}^{(l)}))
\]

Where \( \mathbf{z}_{c,t}^{(0)}=h_{c,t} \) as per a previous equation, and \( l \) belongs to the set \( \{0,1,...,L\} \).

The aggregation function for the \( l \)-th layer, \( g_l(\cdot) \), can be a mean, pooling, or graph convolution (GCN) function. 

In this process, \( \mathbf{a}_{c,t}^{(l)} \) is first concatenated with \( \mathbf{z}_{c,t}^{(l-1)} \), and then transformed using the weight matrix \( \mathbf{W}^{(l)} \). The non-linear function \( \sigma(\cdot) \) is applied to this product to obtain the final embedding for the \( l \)-th layer.
\subsection{Vegetation Indices}
Vegetation indices are essential metrics used in the field of remote sensing phenology to quantify vegetation cover, assess plant health, and (in our study) to estimate crop yields. These indices leverage the spectral data gathered by electromagnetic radiation, which measure various wavelengths of light absorbed and reflected by plants. A fundamental understanding of how these wavelengths interact with vegetation is crucial for interpreting these indices. Specifically, the pigments in plant leaves, such as chlorophyll, absorb wavelengths in the visible spectrum, particularly the red light. Conversely, leaves reflect a significant amount of near-infrared (NIR) light, which is not visible to the human eye. The indices used in the construction of Agri-GNN are available in Appendix \ref{appendix}.

One of the most commonly used vegetation indices is the Normalized Difference Vegetation Index (NDVI). It is calculated using the formula \( \text{NDVI} = \frac{(NIR - RED)}{(NIR + RED)} \), where \( NIR \) represents the near-infrared reflectance and \( RED \) is the reflectance in the red part of the spectrum. The NDVI value ranges between -1 and 1, with higher values typically indicating healthier vegetation and lower values signifying sparse or stressed vegetation. This index is invaluable for various applications, ranging from environmental monitoring to precision agriculture. For Agri-GNN, vegetation indices like NDVI can serve as informative node attributes in agronomic graphs, enhancing the model's ability to make accurate and meaningful predictions in agricultural settings\citep{bannari1995review}.


\section{Methods}
Our proposed framework, \textit{Agri-GNN}, aims to provide a comprehensive solution to crop yield prediction by leveraging the power of Graph Neural Networks (GNNs). The methods section goes over the various stages involved in the design, construction, and validation of \textit{Agri-GNN}.

\subsection{Graph Construction}\label{sec:graph_cons}

To effectively utilize GNNs for agricultural prediction, we first represent the agricultural data as a graph \( \mathcal{G} = (\mathcal{V}, \mathcal{E}) \), where \( \mathcal{V} \) denotes the set of nodes and \( \mathcal{E} \) denotes the set of edges.

\subsubsection{Node Representation}

Each node \( v_i \in \mathcal{V} \) corresponds to a specific plot and is associated with a feature vector \( \mathbf{x}_i \). This feature vector encapsulates all given data information:

\textbf{Vegetation Indices (\( \mathbf{r}_i \))}: The variable \( r_i \) is derived from remote sensing imagery and encapsulates various spectral indices, including but not limited to reflectance values, that are crucial for assessing vegetation health and vitality. These spectral indices often make use of different reflectance bands, such as near-infrared and red bands, to capture detailed information about plant life. These indices serve as a valuable source of information, particularly when integrated with other types of data like genotypic and soil information. The computational procedure to obtain \( r_i \) often involves sophisticated algorithms that account for atmospheric corrections and other sources of noise in the imagery. The specific formulas and methodologies used to determine \( r_i \) are detailed in Appendix \ref{appendix}.

\textbf{Genotypic Data (\( \mathbf{g}_i \))}: The variable \( g_i \) encapsulates the genetic information associated with a specific crop or plant. This genotypic data serves as a foundational element in the realm of Cyber Agricultural Systems. Utilizing advanced imaging techniques like hyperspectral imaging, along with computational methods such as machine learning algorithms, the acquisition and analysis of \( g_i \) have been significantly streamlined. These advancements not only ease the process of data collection but also enable more accurate and comprehensive genetic profiling. Such in-depth genotypic information is invaluable for understanding plant characteristics, disease resistance, and yield potential, thereby playing a crucial role in the development of precision agriculture strategies and sustainable farming practices~\cite{singh2021high}.

\textbf{Weather Data (\( \mathbf{w}_i \))}: The variable \( w_i \) encompasses an array of meteorological factors related to a specific agricultural plot, capturing elements such as temperature, precipitation, humidity, wind speed, and solar radiation. These weather conditions are collected through a variety of methods, including on-site weather stations, remote sensors, and even satellite data. The comprehensive nature of \( w_i \) allows it to serve as a vital input for crop health monitoring systems and predictive yield models. For instance, high-resolution temporal data on factors like soil moisture and air temperature can be instrumental in predicting potential stress events for crops, such as drought or frost risks. Furthermore, when integrated with other data types like genotypic and vegetation indices, \( w_i \) contributes to creating a multifaceted, dynamic model of the agricultural environment(\cite{mansfield2005relationships}).

The final node feature vector is a concatenation of these features:

\begin{equation}
\mathbf{x}_i = [\mathbf{r}_i, \mathbf{g}_i, \mathbf{w}_i]
\end{equation}

\subsubsection{Edge Representation in \textit{Agri-GNN}}

Edges play an indispensable role in graph-based neural network architectures, encoding vital relationships between nodes. Within the \textit{Agri-GNN} architecture, the edge set \( \mathcal{E} \) is meticulously constructed to encapsulate the intricate relationships between agricultural plots. This is achieved by harnessing both spatial and genotypic attributes.

There are two edge constructions that are undertaken. $ \mathcal{E}_{\text{spatial}}$ encompasses the edges that are created through spatial proximity. Given the geographical coordinates \( \mathbf{c}(v_i) \) associated with a node \( v_i \), the pairwise distance to another node \( v_j \) having coordinates \( \mathbf{c}(v_j) \) can be defined as:

\begin{equation}
d(v_i, v_j) = \| \mathbf{c}(v_i) - \mathbf{c}(v_j) \|
\label{eq:distance}
\end{equation}

For every node \( v_i \), edges are constructed to nodes that fall within the bottom 3\% of all pairwise distances, thereby ensuring that the model captures localized environmental intricacies and dependencies:

\begin{equation}
e_{ij} = 
\begin{cases} 
1 & \text{if } d(v_i, v_j) \in \text{bottom 3\% of distances for } v_i \\
0 & \text{otherwise}
\end{cases}
\label{eq:spatial_edge}
\end{equation}

The second set of edges constructed is represented by $ \mathcal{E}_{\text{genotypic}}$. The genotypic data serves as a repository of the genetic characteristics of agricultural plots, offering a window into inherent traits and susceptibilities. Let \( g(v_i) \) represent the genotypic data for node \( v_i \). An edge is formed between nodes \( v_i \) and \( v_j \) if their genotypic attributes resonate:

\begin{equation}
e_{ij} = 
\begin{cases} 
1 & \text{if } g(v_i) = g(v_j) \\
0 & \text{otherwise}
\end{cases}
\label{eq:genotypic_edge}
\end{equation}

The culmination of the edge formation process results in the edge set \( \mathcal{E} \), which is a fusion of edges derived from both spatial proximity and genotypic similarity:

\begin{equation}
\mathcal{E} = \mathcal{E}_{\text{spatial}} \cup \mathcal{E}_{\text{genotypic}}
\label{eq:edge_set}
\end{equation}

By harmonizing spatial and genotypic data, \textit{Agri-GNN} crafts a robust and nuanced representation of the agricultural milieu, establishing itself as an efficacious tool for diverse applications in the realm of agriculture.

\subsection{Graph Neural Network Architecture}

For our crop yield prediction task, we introduce the \textit{Agri-GNN} model, an adaptation of the GraphSAGE (Graph Sample and Aggregation) architecture discussed in section \ref{sec:background}. The model operates on graph \( \mathcal{G} \), designed for efficient aggregation and propagation of information across the graph.

\textit{AgriGNN} has four GraphSAGE convolutional layers to process and refine node features, ensuring an effective representation for downstream tasks. The model architecture is explained in detail in this section.

In the initial layer of the Architecture, whose primary objective is transitioning the input node features to an intermediate representation using hidden channels. The transformation for node \(i\) in this layer is shown by equation \eqref{eq:init_layer}.

\begin{equation}
\mathbf{h}_i^{(1)} = \sigma \left( \mathbf{W}_{\text{init}} \cdot \mathbf{x}_i + \mathbf{b}_{\text{init}} \right)
\label{eq:init_layer}
\end{equation}

In equation \eqref{eq:init_layer}, \( \mathbf{x}_i \) stands for the initial input features of node \( i \), while \( \mathbf{W}_{\text{init}} \) and \( \mathbf{b}_{\text{init}} \) denote the weight matrix and bias vector of this layer, respectively. The function \( \sigma \) represents the activation function. The Rectified Linear Unit (ReLU) is used.

A salient feature of the \textit{Agri-GNN} is its intermediary layers, which not only collate features from neighboring nodes but also incorporate skip connections to preserve the essence of the original node's features. The aggregation of features from neighboring nodes in these layers is depicted in equation \eqref{eq:agg_inter}.

\begin{equation}
\mathbf{m}_i^{(l)} = \text{AGGREGATE} \left( \{\mathbf{h}_j^{(l-1)}\, \forall j \in \text{N}(i) \} \right)
\label{eq:agg_inter}
\end{equation}

Subsequent to this aggregation, the features undergo a transformation, as expressed in equation \eqref{eq:inter_layers}.

\begin{equation}
\mathbf{h}_i^{(l)} = \sigma \left( \mathbf{W}^{(l)} \cdot \text{CONCAT}(\mathbf{h}_i^{(l-1)}, \mathbf{m}_i^{(l)}) + \mathbf{b}^{(l)} \right)
\label{eq:inter_layers}
\end{equation}

Here, \( \text{N}(i) \) represents the neighboring nodes of node \( i \), and \( \mathbf{W}^{(l)} \) and \( \mathbf{b}^{(l)} \) signify the weight matrix and bias vector for layer \( l \), respectively.

The architecture culminates in the final layer, a pivotal component that produces the model's refined output. This layer mirrors the operations of the intermediary layers in aggregating neighboring node features, but distinguishes itself by excluding the addition of original node features. The aggregation of features in this layer is portrayed in equation \eqref{eq:final_agg}.

\begin{equation}
\mathbf{m}_i^{(4)} = \text{AGGREGATE} \left( \{\mathbf{h}_j^{(3)}\, \forall j \in \text{N}(i) \} \right)
\label{eq:final_agg}
\end{equation}

The subsequent transformation, harnessing the aggregated features to yield the final output, is described in equation \eqref{eq:final_layer}.

\begin{equation}
\mathbf{h}_i^{(4)} = \sigma \left( \mathbf{W}^{(4)} \cdot \text{CONCAT}(\mathbf{h}_i^{(3)}, \mathbf{m}_i^{(4)}) + \mathbf{b}^{(4)} \right)
\label{eq:final_layer}
\end{equation}

To ensure stability in convergence and enhance generalization, each hidden layer is succeeded by a batch normalization step. After normalization, dropout regularization with a rate of \( p = 0.5 \) is employed to combat overfitting, as described by equation \eqref{eq:dropout}:

\begin{equation}
\mathbf{h}_i^{(l)} = \text{Dropout}(\mathbf{h}_i^{(l)}, p=0.5)
\label{eq:dropout}
\end{equation}

The final output of the model is a prediction of the yield of the given node(s).

The final \textit{Agri-GNN} model is designed to take in initial node features \( \mathbf{x}_i \) and produce an output \( \mathbf{o}_i \) for each node, representing the predicted yield. The model is summarized as:

\begin{equation}
\mathbf{o}_i = \text{\textit{Agri-GNN}}(\mathbf{x}_i; \mathcal{G}, \Theta)
\end{equation}

Here, \( \Theta \) stands for the set of all learnable parameters within the model.

The model's performance is evaluated using a Mean Squared Error (MSE) loss between the predicted crop yields \( \mathbf{o}_i \) and the actual yields. Optimization is carried out via the Adam optimizer, and hyperparameters such as learning rates are fine-tuned for optimal performance.

\begin{figure}[H]
    \centering
    \includegraphics[width = .75\textwidth]{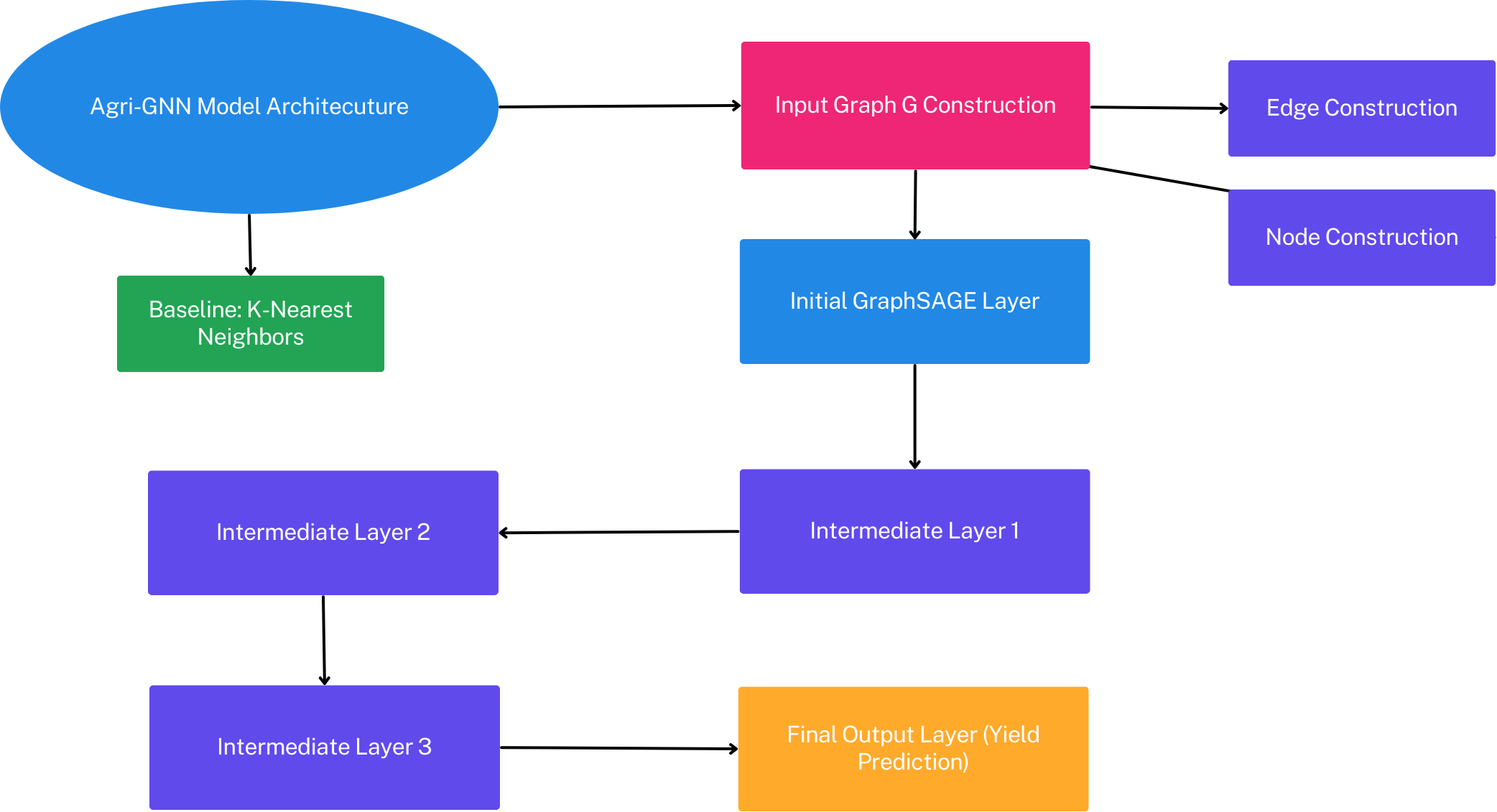}
    \caption{Summary of Model Architecture}
    \label{fig:model-arch}
\end{figure}

\textit{Agri-GNN}'s architecture is summarized in Figure \ref{fig:model-arch}. 

\section{Applications and Experimental Results}
The \textit{Agri-GNN} framework is now applied to plot fields in Ames, Iowa. \textit{Agri-GNN}'s performance is compared to various baseline yield prediction models to accurately gauge its potential.
\subsection{Data Collection and Processing}
\label{data}
Data on soil attributes, hyperspectral reflectance, seed yield, and weather were collected systematically over two consecutive years, 2020 and 2021. The data collection covered both Preliminary Yield Trials (PYT) and Advanced Yield Trials (AYT), and each year involved multiple trial locations.

\subsubsection{Trial Design}
The Preliminary Yield Trials (PYT) were designed using a row-column configuration. In this layout, each plot had a width of 1.52m and a length of 2.13m. Plots within each row were interspaced by 0.91m. The Advanced Yield Trials (AYT) followed a similar row-column design, but with each plot measuring 1.52m in width and 5.18m in length. The interspacing between plots remained consistent at 0.91m for both trial types.

\subsubsection{Soil Data Collection}

Digital soil mapping techniques were used to identify ten specific soil attributes, supplementing the collection of hyperspectral data. Soil cores were extracted down to a depth of 15 cm following a 25m grid sampling pattern, using specialized soil probes. Digital soil maps were then generated with 3mx3m pixel resolution using the Cubist regression machine learning algorithm\citet{khaledian2020selecting}.

For each plot, boundaries were outlined using polygon shape files, and the average value of each soil feature was calculated. To improve data reliability, a 3x3 moving mean was computed for each soil attribute within the plot, and this smoothed value was then used for more detailed analyses. The assessed soil features encompassed Calcium (Ca), Cation Exchange Capacity (CEC), Potassium (K), Magnesium (Mg), Organic Matter (OM), Phosphorus (P1), Percent Hydrogen (Ph), and proportions of Clay, Sand, and Silt.

\subsubsection{Hyperspectral Reflectance Data}

\begin{figure}[H]
    \centering
    \includegraphics[width = .75\textwidth]{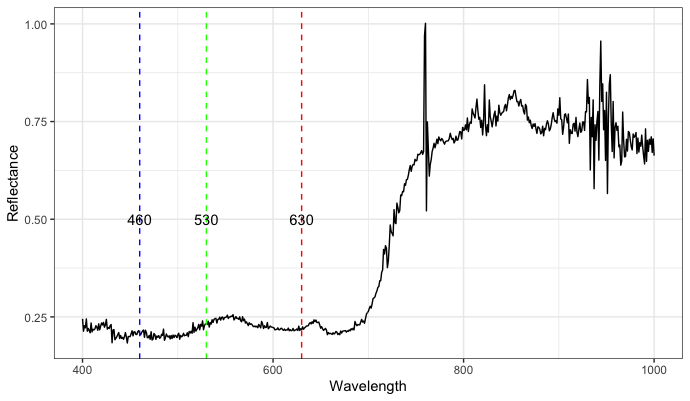}
    \caption{Reflectance plot showcasing wavelengths from 400 nm to 1000 nm. The depicted red, green, and blue bands are illustrative of the visible spectrum wavelengths.}
    \label{fig:reflect}
\end{figure}

Hyperspectral reflectance data for each plot was captured using a Thorlabs CCS200 spectrometer, based in Newton, NJ. The methodology adheres to the system outlined in the study by \cite{bai2016multi}. Spectral data was collected annually at three distinct time points: T1, T2, and T3, captured sequentially, covering wavelengths from 200 nm to 1000 nm, as illustrated in Figure \ref{fig:reflect}.

Particular emphasis is placed on the T3 timepoint, which has been identified as having superior feature importance values according to preliminary data assessments. As vegetation nears physiological maturity, the correlation between hyperspectral reflectance data and crop yield becomes more significant.

Fifty-two vegetation indices were calculated based on the collected hyperspectral reflectance values, following the methodology detailed in the study by \cite{zongpengetal22}. A comprehensive list of these indices is available in Appendix \ref{appendix}. The distribution of the collected data across the four fields is summarized in Table \ref{tab:field_num}.

\begin{table}[H]
    \centering
    \caption{Number of datapoints in each of the four fields.}
    \begin{tabular}{cc}
    \hline\hline\\
    Field No. & Number of observations\\
    \hline\hline
    1 &  770\\
    2 & 912\\
    3 & 800\\
    4 & 679\\
    \hline\hline
    \end{tabular}
    \label{tab:field_num}
\end{table}

\begin{figure}[H]
    \centering
    \includegraphics[width = .75\textwidth]{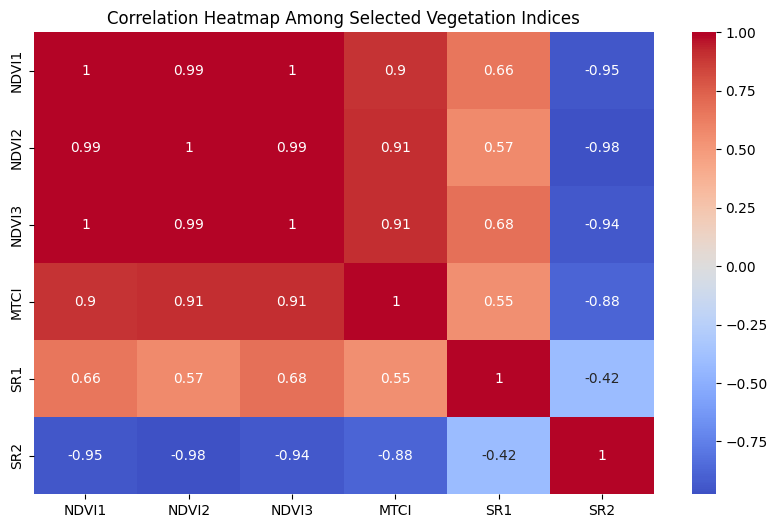}
    \caption{Correlation of selected vegetation indices is shown. }
    \label{fig:corr_vegetation}
\end{figure}

It is interesting to note that many of the vegetation indices show a strong correlation with each other, especially when considering their underlying mathematical formulations. As depicted in Figure \ref{fig:corr_vegetation}, certain pairs of indices exhibit notably high correlation values, suggesting that they might be capturing similar information about the vegetation. This redundancy could be attributed to the fact that many vegetation indices are derived from the same spectral bands, primarily the red and near-infrared (NIR) regions, which are known to be indicative of plant health and vigor. However, while two indices might be highly correlated, they might still provide unique insights to different vegetation properties, and thus, all of the vegetation indices were kept as features in the construction of the dataset.

\subsubsection{Yield Data}
\label{yielddata}

Seed yield data at the plot level was collected using an Almaco small plot combine, headquartered in Nevada, IA. For consistency, all yield data was normalized to a moisture content of 13\% and then converted to kilograms per hectare (Kg/ha). The data collection process was organized in blocks, mirroring the layout established by the breeding program. This arrangement grouped the plots based on their genetic lineage and corresponding maturity groups.

Prior to the computation of vegetation indices, a rigorous data preprocessing phase was undertaken to omit anomalies and outliers. The steps encapsulated:

\begin{enumerate}
    \item Omission of all observations for band values falling below 400 nm due to detected anomalies in these bands' readings.
    \item Exclusion of datapoints with negative hyperspectral values within the 400 nm to 1000 nm band range.
    \item Removal of datapoints showcasing negative seed yield values.
\end{enumerate}

\subsection{Graph Construction}

Once the dataset was pruned to retain only the relevant columns, we started the task of graph construction, as explained in detail in Section \ref{sec:graph_cons}. The nodes of the graph represented individual data points from the dataset, while the edges encoded two primary relationships: spatial proximity and genotype similarity.

The geospatial coordinates (`Latitude` and `Longitude`) of each data point facilitated the computation of pairwise distances between them. By harnessing this distance matrix, we established a threshold—specifically the 3rd percentile of non-zero distances—and used it as a criterion to draw edges between nodes. In essence, if the spatial distance between two nodes was less than this threshold, an edge was drawn between them. 


Beyond spatial relationships, our graph also recognized the significance of genetic similarities between data points. This was achieved by drawing edges between nodes that shared the same genotype, as denoted by the `Population` column. By adopting this strategy, our graph was enriched with edges that encapsulated intrinsic genetic relationships, which has an influence on agricultural yield \citep{shook2021crop}

To facilitate subsequent graph-based deep learning, we represented our graph using the \textit{PyTorch Geometric} framework. This involved a series of transformations. Firstly, the combined spatial and genotype edges were aggregated. Secondly, the dataset was processed to handle missing values, by imputing them with column-wise means, and categorical columns were one-hot encoded. The final graph representation incorporated node features (derived from the processed dataset), the aggregated edges, and the target label (`Yield`).

The resulting graph, thus constructed, served as the foundation for our \textit{\textit{Agri-GNN}} experiments.

\subsection{Neural Architecture}

The model was developed using the \textit{PyTorch} framework \citep{paszke2017automatic}. For the specific needs of graph-based neural networks, we turned to the \textit{PyTorch Geometric (PyG)} library (\cite{fey2019fast}). This library is a comprehensive toolkit, providing a range of architectures and tools tailored for graph neural networks.

Our model, \textit{\textit{Agri-GNN}}, is an augmentation of the conventional GraphSAGE architecture \cite{hamilton2017inductive}. It features four GraphSAGE convolutional layers. The initial layer transitions input features to hidden channels. The two intermediary layers, enhanced with skip connections, amplify the model's capacity to discern both rudimentary and advanced patterns in the data. The final layer is designed to yield the model's output. To ensure the stability and efficiency of training, batch normalization is applied following each convolutional layer. Furthermore, to mitigate the risk of overfitting, dropout regularization is integrated after each batch normalization, with a rate of \(0.5\).

The dimensionality of the input features was determined dynamically based on the dataset. The model was trained for \(500\) epochs, and was monitored to gauge the model's performance accurately and take measures against potential overfitting.

An $80-20$ split was utilized, where $80\%$ of the farm nodes were randomly chosen to be part of the training dataset, 

\subsection{Hyperparameter Tuning}

Hyperparameter tuning was critical to ensuring we have the optimal model. We conducted an exhaustive exploration of various hyperparameter combinations to pinpoint the most conducive setting for our model. The hyperparameters that we varied are summarized in Table \ref{tab:hyperparameters}.

\begin{table}[H]
    \centering
    \begin{tabular}{|c|c|}
        \hline
        \textbf{Hyperparameter} & \textbf{Values Explored} \\
        \hline
        Learning Rates & \(0.001, 0.005, 0.01, 0.02\) \\
        \hline
        Hidden Channels & \(32, 64, 128\) \\
        \hline
        Dropout Rates & \(0.3, 0.5, 0.7\) \\
        \hline
    \end{tabular}
    \caption{Hyperparameters and their explored values}
    \label{tab:hyperparameters}
\end{table}

The best hyperparameters can be seen in Table \ref{tab:best_hyperparameters}.

\begin{table}[H]
    \centering
    \begin{tabular}{|c|c|}
        \hline
        \textbf{Hyperparameter} & \textbf{Best Value} \\
        \hline
        Learning Rate & \(0.02\) \\
        \hline
        Hidden Channels & \(32\) \\
        \hline
        Dropout Rate & \(0.3\) \\
        \hline
    \end{tabular}
    \caption{Optimal hyperparameters after tuning}
    \label{tab:best_hyperparameters}
\end{table}
\section{Model Performance and Results}

We now show the results of \textit{\textit{Agri-GNN}} on the described dataset. The results are compared with two baselines on the same dataset: a K-Nearest Neighbors Model and an Ensemble Machine Learning model~\citep{chattopadhyay2023comprehensive}.

\begin{figure}[H]
    \centering
    \includegraphics[width = .75\textwidth]{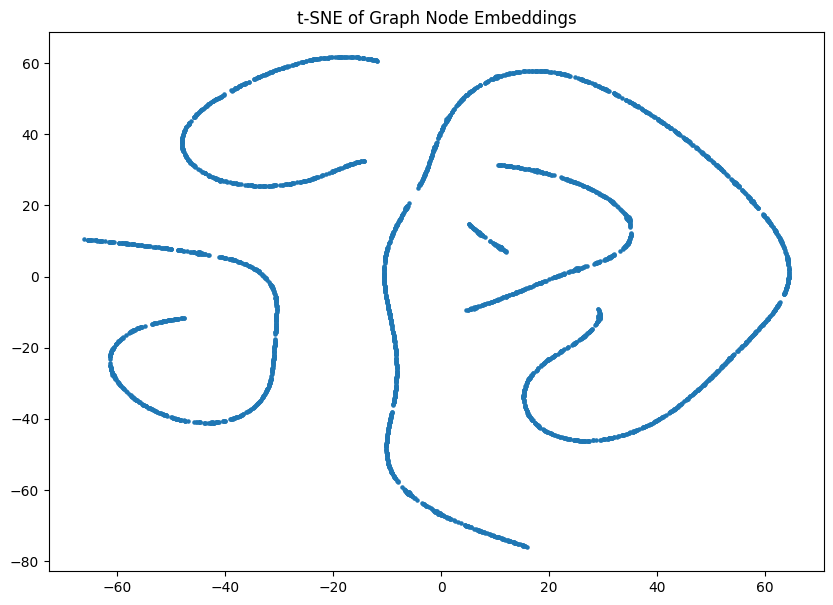}
    \caption{t-SNE visualization of graph node embeddings.}
    \label{fig:tsne-embedding}
\end{figure}

We first visualize the embeddings derived from our graph neural network model. To better understand the spatial distribution of these embeddings, we employed the t-Distributed Stochastic Neighbor Embedding (t-SNE) algorithm for dimensionality reduction (\cite{hinton2002stochastic}). The embeddings, initially generated by the forward method of our model, were reduced to two dimensions using t-SNE. As shown in Figure \ref{fig:tsne-embedding}, each point represents a node in the graph, and the spatial arrangement captures the similarity between node embeddings. Distinct clusters and patterns in the visualization indicate nodes with similar features within the graph. The embeddings can be seen in Figure \ref{fig:tsne-embedding}.

From Figure \ref{fig:tsne-embedding}, we see that multiple distinct graphs may possibly be formed from the data, particularly when subsets of the data are not related in any spatial or genotypic way. \textit{Agri-GNN} is designed to be able to learn the inherent patterns in each of these subgraphs, while further increasing the accuracy of the main base model. 

The performance of \textit{\textit{Agri-GNN}} was assessed using standard regression metrics, as presented in table \ref{tab:performance_metrics}:

\begin{table}[h]
    \centering
    \begin{tabular}{|c|c|}
        \hline
        \textbf{Metric} & \textbf{\textit{Agri-GNN} Value} \\
        \hline
        Root Mean Squared Error (RMSE) & \(4.565\) \\
        \hline
        Mean Absolute Error (MAE) & \(3.590\) \\
        \hline
        Coefficient of Determination (\(R^2\)) & \(0.876\) \\
        \hline
    \end{tabular}
    \caption{Performance metrics of the \textit{Agri-GNN}} model
    \label{tab:performance_metrics}
\end{table}

\textit{Agri-GNN} achieves a Root Mean Squared Error (RMSE) of \(4.565\) on this dataset. The Mean Absolute Error (MAE) is \(3.590\). The model has an \(R^2\) value of \(0.876\). These results show significant improvement over the baselines used.

The K-Nearest Neighbor (K-NN) Algorithm is used as the first baseline model. The K-NN model predicts the yield of a node based on the average yield of its \(k\) nearest neighbors based on latitude and longitude. This model is optimized by performing a grid search to find the best value of \(k\) from a range of 1 to 20. The optimized K-NN model used \(k = 18\) as the number of neighbors for making predictions.

The performance metrics of the optimized K-NN model are presented in Table \ref{tab:baseline_metrics}. The model achieved a Root Mean Squared Error (RMSE) of \(12.93\), a Mean Absolute Error (MAE) of \(10.33\), and a Coefficient of Determination (\(R^2\)) of \(0.026\). The \textit{Agri-GNN} model shows superior performance to the baseline. The metrics signify a substantial enhancement over the baseline K-NN model, demonstrating the effectiveness of \textit{Agri-GNN} in yield prediction.

\begin{table}[h]
    \centering
    \begin{tabular}{|c|c|}
        \hline
        \textbf{Metric} & \textbf{\textit{Baseline K-NN} Value} \\
        \hline
        Root Mean Squared Error (RMSE) & \(12.93\) \\
        \hline
        Mean Absolute Error (MAE) & \(10.33\) \\
        \hline
        Coefficient of Determination (\(R^2\)) & \(0.026\) \\
        \hline
    \end{tabular}
    \caption{Performance metrics of K-NN baseline model ($k=18$}
    \label{tab:baseline_metrics}
\end{table}

\textit{Agri-GNN}'s prediction error and $R^2$ shows significant potential when compared to recent work on a similar dataset~\citep{chattopadhyay2023comprehensive}, where the highest $R^2$ value achieved was less than $.8$. Such an enhancement underscores the potential benefits of graph-based neural networks, especially when handling datasets with inherent relational structures.

\begin{figure}[H]
    \centering
    \includegraphics[width = .65\textwidth]{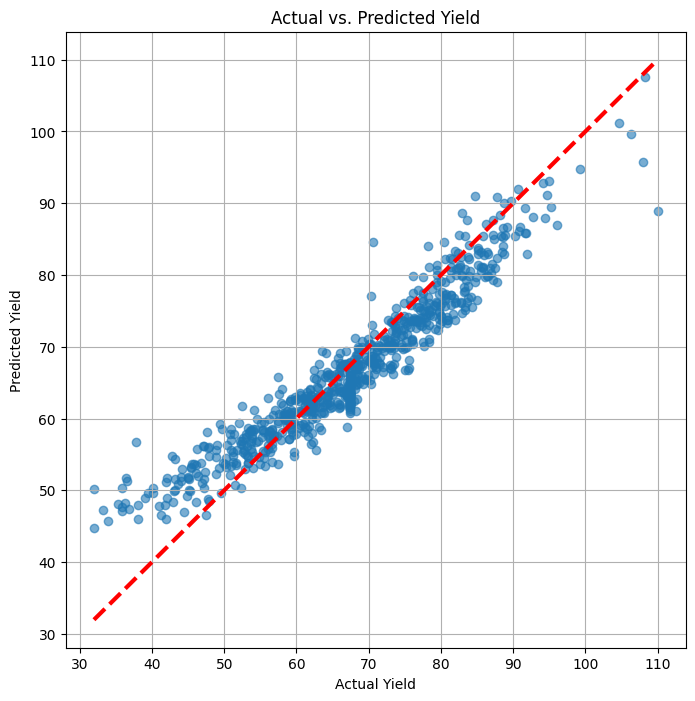}
    \caption{The Actual vs. Predicted Yield of our predictions. The red line signifies a perfect model. The model achieves an \(R^2\) value of \(0.876\) }
    \label{fig:actual-predicted}
\end{figure}

In Figure \ref{fig:actual-predicted}, we present a comparison between the Actual and Predicted Yields. This graphical representation provides a comprehensive insight into the accuracy and precision of our model's predictions. An interesting observation from Figure \ref{fig:actual-predicted} is the remarkable accuracy of the model's predictions for yields ranging between 50 and 90. The data points in this range cluster closely around the line of perfect agreement, suggesting that the model is particularly adept at predicting yields within this interval. This is noteworthy as it indicates that our model is not only reliable in general but also exhibits enhanced performance when predicting yields in this specific range.

\subsection{Scalability and Robustness}

In the previous section, an application of \textit{Agri-GNN} is outlined, focusing on agricultural yield prediction in farms located in Ames, Iowa. The results demonstrate the model's capability to efficiently and accurately predict yields based on geographical coordinates. This section delves into the scalability and robustness of the \textit{Agri-GNN} framework, discussing its potential application on a broader scale, both nationally and internationally.

\textit{Agri-GNN} is designed with scalability in mind, allowing for seamless integration and deployment in diverse farming contexts across the world. The model's adaptability to various geographical and environmental conditions ensures consistent and reliable yield predictions irrespective of the location. The use of cloud computing resources and distributed computing frameworks enhances the scalability of \textit{Agri-GNN}, enabling real-time processing and analysis of large datasets encompassing numerous farms across different geographical locations.

Robustness is at the core of the \textit{Agri-GNN} framework. The model employs advanced machine learning algorithms and techniques to ensure stability and reliability in yield predictions, even in the face of data variability and uncertainties. The robust nature of \textit{Agri-GNN} ensures uninterrupted and consistent performance, bolstering the confidence of farmers and stakeholders in the accuracy and reliability of the yield predictions provided by the model. Moreover, the continuous learning and adaptation capabilities of \textit{Agri-GNN} further enhance its robustness, ensuring it remains at the forefront of agricultural yield prediction technology. Unfortunately, \textit{Agri-GNN} is optimized to work best with large datasets where ample information from previous experiments is available. In many farming contexts, there is not enough information to be able to train $Agri-GNN$ to have reliable performance~\citep{wiseman2019farmers}. Further research should explore how \textit{Agri-GNN} can be optimized to perform well in instances where a lack of ample farming data may be a problem.



\section{Conclusion}\label{sec:conclusion}

In an ever-evolving agricultural landscape marked by climatic uncertainties, the pressing need for accurate and holistic crop yield predictions has never been greater. This study introduced Agri-GNN, a pioneering approach that harnesses the power of Graph Neural Networks to provide a comprehensive representation of the agricultural ecosystem. Unlike traditional models that often operate in silos, Agri-GNN's strength lies in its ability to synthesize diverse data modalities into a unified framework, capturing the intricate interplay between genetic, environmental, and spatial factors.

Agri-GNN transcends conventional methodologies by viewing the agricultural ecosystem as an interconnected network, where crops aren't just passive entities but active nodes influenced by their immediate surroundings and broader contexts. This perspective, combined with GNN's superior data processing capabilities, enables Agri-GNN to deliver predictions that are both granular and holistic.

Furthermore, Agri-GNN's modular design ensures its relevance in a rapidly changing agricultural sector, allowing for seamless integration of new data sources and insights. Its precision agriculture approach not only aids in enhancing productivity but also paves the way for sustainable practices that respect both economic and environmental considerations.

Our applications of the Agri-GNN framework's capabilities to the plot fields of Ames, Iowa show promising results, even obtaining better performance than the results obtained in \cite{chattopadhyay2023comprehensive}. Agri-GNN's performance metrics, including $RMSE$, $MAE$, and 
$R^2$ highlighted its proficiency in yield prediction. Notably, the model showcased significant improvements over existing models, reaffirming the potential of graph-based neural networks in agricultural applications. The t-SNE visualizations further provided insights into the model's embeddings, reinforcing the cohesive and interconnected nature of the data.

In summary, Agri-GNN represents a paradigm shift in agricultural modeling. By capturing the complexity and interconnectedness of farming systems, it offers a fresh lens through which to view and address the multifaceted challenges of modern agriculture. As we stand at the crossroads of traditional practices and technological innovation, Agri-GNN serves as a beacon, guiding the way towards informed, sustainable, and resilient agricultural futures.

\section{Acknowledgments}

The authors thank Joscif Raigne, Dr. Baskar Ganapathysubramanian, and Dr. Soumik Sarkar for their invaluable feedback on the draft manuscript.

The authors thank Joscif Raigne for his help in the construction of spatial data for the field experiments.

The authors thank Shannon Denna and Christopher Grattoni for their support.

The authors thank staff and student members of SinghSoybean group at Iowa State University, particularly Brian Scott, Will Doepke, Jennifer Hicks, Ryan Dunn, and Sam Blair for their assistance with field experiments and phenotyping. 

This work was supported by the Iowa Soybean Association, North Central Soybean Research Program, USDA CRIS project IOW04714, AI Institute for Resilient Agriculture (USDA-NIFA \#2021-647021-35329), COALESCE: COntext Aware LEarning for Sustainable CybEr-Agricultural Systems (CPS Frontier \#1954556), Smart Integrated Farm Network for Rural Agricultural Communities (SIRAC) (NSF S \& CC \#1952045), RF Baker Center for Plant Breeding, and Plant Sciences Institute.

\section{Author Contributions}
Aditya Gupta conceptualized and designed the proposed model, developed the methodology, conducted the data analysis, constructed and implemented the Agri-GNN, provided visualization of the results, and took primary responsibility for writing, revising, and finalizing this paper. All authors have read, reviewed, and agreed to the published version of the manuscript.
\section{Conflict of Interest}
The authors declare no conflict of interest.

\bibliographystyle{plainnat}
\bibliography{ref}

\appendix
\newpage
\section{Table of Vegetation Indices}

\label{appendix}
\noindent
\begin{table*}[h]
    \noindent
    \small
    \caption{Summary of the 52 vegetation indices. Here Tx denote the hyperspectral reflectance value at x nm.}
    \resizebox{15cm}{!}{
    \begin{tabular}{lll}
    \hline\hline\
    Full form & Spectral Index/Ratio & Formula\\
    \hline\hline
    Curvature index & Cl & $T675 \times T690/T683^2$\\
    Chlorophyll Index red-edge & Clre & $T750/T710 - 1$\\
     & Datt1 & $(T850 - T710)/(T850 - T680)$\\
     & Datt4 & $T672/(T550 \times T708)$\\
     & Datt6 & $T860/(T550 \times T708)$\\
    Double difference index & DDI & $(T749 - T720) - (T701 - T672)$\\
    Double peak index & DPI & $(T688 + T710)/T697^2$\\
    Gitelson2 & & $(T750 - T800)/(T695 - T740) -1$\\
    Green normalized difference vegetation index & GNDVI & $(T750 - T550)/(T750 + T550)$\\
    Modified chlorphyll absorption ratio index & MCARI & $[(T700 - T670) - 0.2(T700 - T550)](T700/T670)$\\
     & MCARI3 & $[(T750 - T710) - 0.2(T750 - T550)](T750/T715)$\\
    Modified normalized difference & MND1 & $(T800 - T680)/(T800 + T680 - 2 \times T445)$ \\
    & MND2 & $(T750 - T705)/(T750 + T705 - 2 \times T445)$ \\
    Modified simple ratio & mSR & $(T800 - T445)/(T680 - T445)$\\
    Modified simple ratio 2 & mSR2 & $(T750/T705 - 1)/(\sqrt{T750/T705 + 1})$\\
    MERIS terrestrail cholrophyll index & MTCI & $(T754 - T709)/(T709 - T681)$\\
    Modified traingular vegetation index 1 & MTVI1 & $1.2[1.2(T800 - T550) - 2.5(T670 - T550)]$\\
    Normalized difference 550/531 & ND1 & $(T550 - T531)/(T550 + T531)$\\
    Normalized difference 682/553 & ND2 & $(T682 - T553)/(T682 + T553)$\\
    Normalized difference chlorophyll & NDchl & $(T925 - T710)/(T925 + T710)$\\
    Normalized difference red edge & NDRE & $(T790 - T720)/(T790 + T720)$\\
    Normalized difference vegetation index & NDVI1 & $(T750 - T650)/(T750 + T650)$\\
     & NDVI2 & $(T750 - T550)/(T750 + T550)$\\
     & NDVI3 & $(T750 - T710)/(T750 + T710)$\\
    Normalized pigment cholrophyll index & NPCL & $(T680 - T430)/(T680 + T430)$\\
    Normalized difference pigment index & NPQI & $(T415 - T435)/(T415 + T435)$\\
    Optimized soil-adjusted vegetation index & OSAVI & $(1 + 0.16)(T800 - T670)(T800 + T670 - 0.16)$\\
    Plant biochemical index & PBI & $T810/T560$\\
    Plant pigment ratio & PPR & $(T550 - T450)/(T550 + T450)$\\
    Physiological reference index & PRI & $(T550 - T530)/(T550 + T530)$\\
    Pigment-specific normalized difference & PSNDb1 & $(T800 - T650)/(T800 + T650)$\\
     & PSNDc1 & $(T800 - T500)/(T800 + T500)$\\
     & PSNDc2 & $(T800 - T470)/(T800 + T470)$\\
    Plant senescence reflectance index & PSRI & $(T678 - T500)/T750$\\
    Pigment-specific simple ratio & PSSRc1 & $T[800]/T[500]$\\
     & PSSRc2 & $T[800]/T[740]$\\
    Photosynthetic vigor ratio & PVR & $T([550] - T[650])/(T[550] + T[650])$\\
    Plant water index & PWI & $T970/T900$\\
    Renormalized difference vegetation index & RDVI & $(T800 - T670)/\sqrt{(T800 + T670)}$\\
    Red-edge stress vegatation index & RVSI & $((T718 + T748)/2) - T733$\\
    Soil-adjusted vegatation index & SAVI & $1.16((T800 - T670)/(T800 + T670 + 0.16))$\\
    Structure intensive pigment index & SIPI & $(T800 - T445)/(T800 + T680)$\\
    Simple ratio & SR1 & $T430/T680$\\
     & SR2 & $T440/T740$\\
     & SR3 & $T550/T672$\\
     & SR4 & $T550/T750$\\
    Disease -water stress index 4 & DSWI-4 & $T550/T680$\\
    Simple ratio pigment index & SRPI & $T430/T680$\\
    Transformed chlorophyll absorption ratio & TCARI & $3((T700 - T670) - 0.2(T700 - T550)(T700/T670))$\\
    Traingular cholrophyll index & TCI & $1.2(T700-T550) - 1.5(T670 - T550) \times \sqrt{T700/T670}$\\
    Triangular vegetation index & TVI & $0.5(120(T750 - T550) - 200(T670 - T550))$\\
    Water band index & WBI & $T970/T902$\\
    \hline\hline
    \end{tabular}
    }
    \label{tab:health_index}
\end{table*}
\end{document}